\documentclass[11pt,a4paper]{article}
\usepackage[hyperref]{acl2020}

\usepackage{times}
\usepackage{latexsym}
\usepackage{graphicx}

\usepackage{microtype}
\usepackage{amsmath,amssymb,amsfonts}
\usepackage[mathscr]{euscript}
\usepackage{amsbsy}
\usepackage{multirow}
\usepackage{booktabs}
\usepackage{subcaption}
\usepackage{commath}
\usepackage{newtxtext,newtxmath}
\usepackage{relsize}
\usepackage{dsfont}

\usepackage{hyperref}
\usepackage{array,makecell}
\usepackage{microtype}
\usepackage{amsmath,amssymb,amsfonts}
\usepackage[mathscr]{euscript}
\usepackage{amsbsy}
\usepackage{multirow}
\usepackage{booktabs}
\usepackage{subcaption}
\usepackage{commath}
\usepackage{newtxtext,newtxmath}
\usepackage{relsize}
\usepackage{dsfont}
\usepackage{graphicx}
\usepackage{booktabs,chemformula}
\usepackage{mathtools}
\usepackage{bm}
\usepackage{esvect}

\usepackage{times}
\usepackage{latexsym}

\usepackage{microtype}

\usepackage{xcolor}
\newcommand\crule[3][black]{\textcolor{#1}{\rule[-0.5ex]{#2}{#3}}}

\usepackage{tikz}
\usetikzlibrary{fadings}

\definecolor{dirty_green}{rgb}{0.4, 0.494118, 0.172549} 

\definecolor{t1_r1_c1}{rgb}{0.2845, 0.7257, 0.2592} 
\definecolor{t1_r1_c2}{rgb}{0.6101, 0.4945, 0.2044} 

\definecolor{t1_r2_c1}{rgb}{0.5528, 0.2047, 0.6787} 
\definecolor{t1_r2_c2}{rgb}{0.4906, 0.5243, 0.3785} 

\definecolor{t1_r3_c1}{rgb}{0.5528, 0.2047, 0.6787} 
\definecolor{t1_r3_c2}{rgb}{0.6753, 0.1028, 0.1925} 
\aclfinalcopy % Uncomment this line for the final submission
\title{Pragmatically Informative Color Generation \\ by Grounding Contextual Modifiers}

\author{Zhengxuan Wu$^{1}$, Desmond C. Ong$^{2,3}$ \\
$^{1}$Symbolic Systems Program, Stanford University \\
$^{2}$Institute of High Performance Computing, Agency for Science, Technology and Research, Singapore \\
$^{3}$Department of Information Systems and Analytics, National University of Singapore \\
{\tt wuzhengx@stanford.edu, dco@comp.nus.edu.sg} }

\date{}

\begin{document}
\maketitle
\begin{abstract}
Grounding language in contextual information is crucial for fine-grained natural language understanding. One important task that involves grounding contextual modifiers is color generation. Given a reference color ``\emph{green}'', and a modifier ``\emph{bluey}'', how does one generate a color that could represent ``\emph{bluey green}''? We propose a computational pragmatics model that formulates this color generation task as a recursive game between speakers and listeners. In our model, a pragmatic speaker reasons about the inferences that a listener would make, and thus generates a modified color that is maximally informative to help the listener recover the original referents. In this paper, we show that incorporating pragmatic information provides significant improvements in performance compared with other state-of-the-art deep learning models where pragmatic inference and flexibility in representing colors from a large continuous space are lacking. Our model has an absolute 98\% increase in performance for the test cases where the reference colors are unseen during training, and an absolute 40\% increase in performance for the test cases where both the reference colors and the modifiers are unseen during training.\footnote{Code is avaliable at \url{https://github.com/frankaging/Pragmatic-Color-Generation}}

 % Given an reference color and a modifier, a pragmatic speaker generates modified colors that a listener can use to correctly reconstruct the original reference color.

% Although similar techniques have been successfully applied in cognitive science and computational linguistics reference-game tasks (e.g. learning of grounded contextual dependence in reference games), such pragmatic modeling has not been applied to tasks such as color generation, where one has to ground contextual modifiers to generate a sample . 
\end{abstract}

% \section{Credits}

% This document has been adapted
% by Steven Bethard, Ryan Cotterrell and Rui Yan
% from the instructions for earlier ACL and NAACL proceedings, including those for 
% ACL 2019 by Douwe Kiela and Ivan Vuli\'{c},
% NAACL 2019 by Stephanie Lukin and Alla Roskovskaya, 
% ACL 2018 by Shay Cohen, Kevin Gimpel, and Wei Lu, 
% NAACL 2018 by Margaret Michell and Stephanie Lukin,
% 2017/2018 (NA)ACL bibtex suggestions from Jason Eisner,
% ACL 2017 by Dan Gildea and Min-Yen Kan, 
% NAACL 2017 by Margaret Mitchell, 
% ACL 2012 by Maggie Li and Michael White, 
% ACL 2010 by Jing-Shing Chang and Philipp Koehn, 
% ACL 2008 by Johanna D. Moore, Simone Teufel, James Allan, and Sadaoki Furui, 
% ACL 2005 by Hwee Tou Ng and Kemal Oflazer, 
% ACL 2002 by Eugene Charniak and Dekang Lin, 
% and earlier ACL and EACL formats written by several people, including
% John Chen, Henry S. Thompson and Donald Walker.
% Additional elements were taken from the formatting instructions of the \emph{International Joint Conference on Artificial Intelligence} and the \emph{Conference on Computer Vision and Pattern Recognition}.

\section{Introduction}
When describing colors, people rely on comparative adjectives such as ``\emph{pale}'', or ``\emph{bright}''~\cite{lassiter2017adjectival}. Understanding how these comparative words modify the referent color, it requires extensive language grounding, in this case in color space~\cite{monroe2017colors}.
%These comparative words modify colors, which requires extensive language grounding~\cite{monroe2017colors}. 
For instance, understanding (i.e., correctly generating) a color that corresponds to ``\emph{pale} \emph{yellow}'' requires some knowledge about the meaning of ``\emph{yellow}'' and ``\emph{pale}'' in color space, and how the latter modifier modifies the former referent color. This is a difficult and underspecified task, because modifiers like ``\emph{pale}'' are vague, and the strength and effect of a modifier may also depend on the color it is modifying. Understanding such modifiers and how they are grounded in image space is imperative for fine-grained attribute learning in many computer vision tasks~\cite{farhadi2009describing, russakovsky2010attribute, vedaldi2014understanding}.

\begin{figure}[tb]
\centering
\includegraphics[width=1\columnwidth]{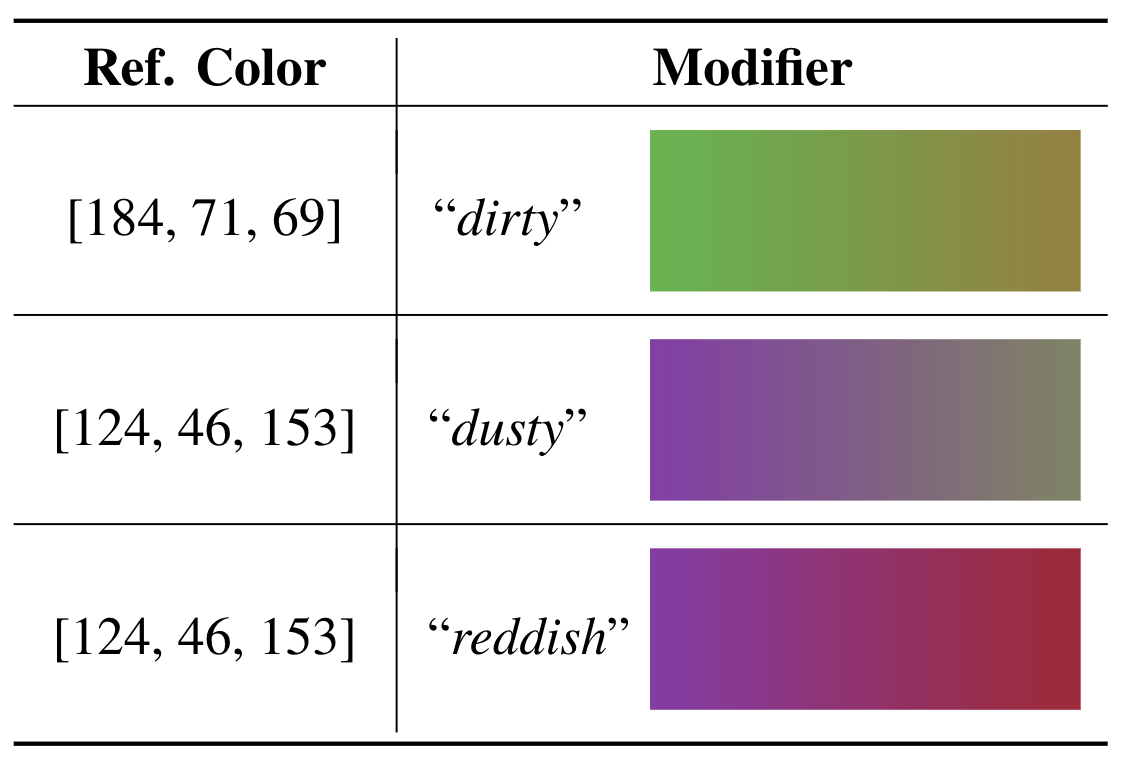}
  \caption{Examples of the color modification task, shown in RGB space. Given the reference color (Ref. Color), the modifier changes the color towards a target color. We represent this modification by going from left to right of the colored bar; the target color is shown at the right-end of the bar. The reference color for the top row is \emph{green}, and \emph{purple} for the middle and bottom rows.}
  \label{table:color}
\end{figure}

% \begin{table}
%   \begin{center}
%     \label{tab:table1}
%     \begin{tabular}{
%     >{\centering\arraybackslash}m{0.3\columnwidth}% instead of "p" is "m"
%     |>{\centering\arraybackslash}m{0.15\columnwidth}
%     >{\centering\arraybackslash}m{0.4\columnwidth}
%     }
%      \toprule
%       \textbf{Ref. Color} & \multicolumn{2}{c}{\textbf{Modifier}}\\
%       \hline \\ [-1.8ex]
%       [184, 71, 69] & ``\textit{dirty}'' & \tikzrule[left color=t1_r1_c1, right color=t1_r1_c2, anchor=east]{0.4\columnwidth}{3em} \\
%       \hline \\ [-1.8ex]
%       [124, 46, 153] & ``\textit{dusty}'' & \tikzrule[left color=t1_r2_c1, right color=t1_r2_c2, anchor=east]{0.4\columnwidth}{3em} \\
%       \hline \\ [-1.8ex]
%       [124, 46, 153] & ``\textit{reddish}'' & \tikzrule[left color=t1_r3_c1, right color=t1_r3_c2, anchor=east]{0.4\columnwidth}{3em} \\
%       \bottomrule
%     \end{tabular}
%   \end{center}
%   \caption{Examples of the color modification task, shown in RGB space. Given the reference color (Ref. Color), the modifier changes the color towards a target color. We represent this modification by going from left to right of the colored bar; the target color is shown at the right-end of the bar. The reference color for the top row is \emph{green}, and \emph{purple} for the middle and bottom rows.}
%   \label{table:color}
% \end{table}

Color provides a simple and tractable space in which to study natural language grounding. 
\citet{monroe2017colors} found that when asked to describe a reference color presented with similar distractors, speakers often rely on comparative adjectives to describe it, rather than relying on context-independent descriptions. 
Others have also studied contextual dependence and vagueness in human-generated color descriptions~\cite{egre2013vagueness, mcmahan2015bayesian}. 
In contrast to these papers, \citet{winn2018lighter} proposed a new paradigm focusing on using machine learning to produce grounded comparative adjectives in color description. As shown in Fig.~\ref{table:color}, given a color ``\emph{green}'' represented by the RGB vector $[184, 71, 69]$ and a modifier ``\emph{dirty}'', the task is to generate RGB vectors for the color ``\emph{dirty green}''. Previous studies have trained deep learning models to generate modified colors by grounding comparative adjectives~\cite{winn2018lighter, han2019grounding}, which produce strong results, but these models lack a notion of pragmatics, which is important for modelling tasks that require extensive language grounding. In parallel, all previous works admit an important limitation where they ignore the richness of the color representation space by only considering a single RGB vector to represent a color label, which is not satisfying as a color label may be represented by a region in the color space.

%This yields strong but still imperfect results, as shown in Fig.~\ref{fig:error}.

In this paper, we propose adding \emph{pragmatic informativeness} with leveraging with the richness of the color representations, a feature present in human contextualized color generation process, but which has been lacking in deep learning approaches to color generation tasks.
%In this paper, we propose a pragmatic approach brings promising improvements in this task by adding \emph{pragmatic informativeness} that is lacking in traditional deep learning approaches. 
The pragmatic approaches have been successfully applied to cognitive science and computational linguistics tasks such as reference games, where one has to understand pragmatics grounded in local context in order to correctly select the target in the presence of distractors~\cite{andreas2016reasoning, monroe2017colors, cohn2018pragmatically, cohn2018incremental, nie2020pragmatic}. These models often build off the Rational Speech Acts (RSA) framework~\cite{frank2012predicting, goodman2013knowledge}, which models a nested reasoning process: a listener reasoning about a speaker reasoning about a listener. Here, we show that pragmatic inference can be similarly applied to grounding contextual information in a \emph{generation} task, specifically color generation in the context of comparative color modification. 

We propose a \emph{reconstructor-based pragmatic speaker} model that learns color generation via pragmatically grounding comparative modifiers. Our work differs from most RSA-based models because we do not have a choice of distractors and referent targets: Rather, the task is to generate a target. Given a reference color (a vector in RGB color space) and a comparative modifier (e.g., ``dirty''), a \emph{literal speaker} model first generates modified colors using deep learning models as in~\cite{winn2018lighter, han2019grounding}. We kept our literal speaker close to previous, state-of-the-art deep learning models that learn a nonlinear mapping from words to (changes in) color space. Building on the literal speaker, we propose a listener that reasons about the literal speaker and tries to guess the original referent color. Finally, our \emph{reconstructor-based pragmatic speaker} reasons about the listener, and produces modified colors that are maximally informative, which would help the listener correctly recover the referent. We find that our pragmatic speaker model performs significantly better than the state-of-the-art models on a variety of test scenarios.

\section{Related Work}

Previous works on comparative color modification---the same task we study here---often rely on deep learning models, such as standard multilayer perceptrons~\cite{winn2018lighter, han2019grounding}, in which pragmatic reasoning is lacking. Additionally, previous works often rely on a single RGB vector representation for a color label when generate colors, which is not convincing as a color label may be represented by a wide range of area in the color representation space~\cite{monroe2017colors}.
Our study serves as an extension on this topic by extending these deep learning models using a pragmatic modeling framework with taking into consideration the richness of color representations.

Pragmatic modeling have been mainly applied to grounded language learning problems in the context of reference games ~\cite{andreas2016reasoning, cohn2018pragmatically, cohn2018incremental, monroe2017colors, nie2020pragmatic} where the model is tasked to distinguish a referent target from a set of similar distractors. These are often based off the RSA framework. Most relevant to the current work, \citet{monroe2017colors} modelled a color reference game where people had to describe a given color (e.g., ``\emph{cyan}'' \crule[cyan]{0.4cm}{0.4cm}) in the presence of distractors, which may be close (e.g., ``\emph{teal}'' \crule[teal]{0.4cm}{0.4cm}) or far (e.g., ``\emph{brown}'' \crule[brown]{0.4cm}{0.4cm}). The authors augmented a deep learning model with pragmatic reasoning to produce a pragmatic listener, which reasons about the speaker producing the color description. Their pragmatic listener more accurately classified correct referents. In contrast to these discrimination problems involving a forced choice selection, our work here focuses on color production, given a reference color and modifier.  

We are aware of only a few recent works that have extended pragmatic modelling to a contextualized generation task, such as ~\citet{fried2017unified} who extended pragmatic modeling to generating and following instructions (e.g. ``walk along the blue carpet''), and ~\citet{shen2019pragmatically}, who considered abstract summarization, which produces a paragraph (e.g., restaurant review) that summarizes a structured input (restaurant type, cuisine, price, customer ratings). 
%While a few recent works have extended pragmatic modelings to traditional natural language understanding tasks such as abstract summarization~\cite{shen2019pragmatically}
Indeed, we think that there is much promise in using these approaches for generative tasks. In fact, our task has the additional complexity of generating images (colors) from language.
%To our best knowledge, pragmatic modeling has not been fully examined in traditional image generation tasks such as color generation. 
%By developing a new pragmatic approach of improving contextual color grounding, our work further proves effectiveness of pragmatic modeling in a simple yet important task of contextual color generation, which suggests potentials in applying such approach in other image generation tasks.

%Though grounding for colors with contextual descriptions are well studied~\cite{monroe2017colors}, there are limited studies explicitly examine comparative color modifications~\cite{winn2018lighter, han2019grounding}, and comparative ranking analysis on image features~\cite{parikh2011relative, yu2014fine}. 

%Previous studies on comparative color modifications often rely on deep learning models---standard multilayer perceptrons~\cite{winn2018lighter, han2019grounding}, in which pragmatic modeling is lacking. 

%\fcolorbox{black}{green}{\rule{0pt}{6pt}\rule{6pt}{0pt}} XXXX

%While a few recent works have extended pragmatic modelings to traditional natural language understanding tasks such as abstract summarization~\cite{shen2019pragmatically}, its main application remains in grounded language learning problems in the context of reference games~\cite{andreas2016reasoning, monroe2017colors, cohn2018pragmatically, cohn2018incremental, nie2020pragmatic} where the model is asked to distinguish a referent target from a set of similar distractors. 

\section{Data} \label{Sec:data}
%\dco{desmond to revisit this paragraph} \zen{Re: Rewrited this section.}
We use the dataset\footnote{\url{https://bitbucket.org/o_winn/comparative_colors}} generated by~\citet{winn2018lighter}. This dataset was initially based on results of a color description survey collected by~\citet{munroe2010color} in which participants were asked to provide free-form labels for colors, giving a collection of mappings between color labels and colors. As a result, this alludes the fact that a single color label may be represented by different colors. This initial survey was then cleaned and filtered by~\citet{mcmahan2015bayesian}, producing a set of 821 unique color labels, with an average of 600 RGB vectors per label. Note that color labels may contain adjective modifiers, such as ``\emph{dirty green}''. \citet{winn2018lighter} converted these mappings to triples of (reference color label, modifier, target color label), such as (``\emph{green}'', ``\emph{dirty}'', ``\emph{dirty green}''), where both the reference color labels and target color labels are included in the original dataset. 
% Note that we expect the modifier describes color transition from a reference color label to a target color label as shown in Table~\ref{table:color}. 
The dataset we use, from ~\citet{winn2018lighter}, contains 415 triples, which includes 79 unique reference color labels and 81 unique modifiers.

We partition the datasets into different sets includes a \emph{Train} set, a \emph{Validation} set and a collection of \emph{Test} sets. The \emph{Train} set (271 triples) and the \emph{Test} sets (415 triples, described in Sec.~\ref{sec:eval}) are partitioned as in the original paper~\cite{winn2018lighter}. The \emph{Validation} set (100 triples) contains triples sampled from the \emph{Train} set for hyper-parameter tuning (not used for training). Additionally, RGB vectors of a color are partitioned across these three sets so that different sets use distinct RGB vectors to represent a color label; thus, if a color label (e.g. ``\emph{red}'') appears as a reference color in both training and testing sets, we will use different sets of RGB vectors for them. We use the same approach presented in~\citet{winn2018lighter}: using the mean value of a set of RGB vectors to represent the gold-standard RGB vector for a target color label for performance comparison.

\begin{figure}[t]
\centering
\includegraphics[width=1\columnwidth]{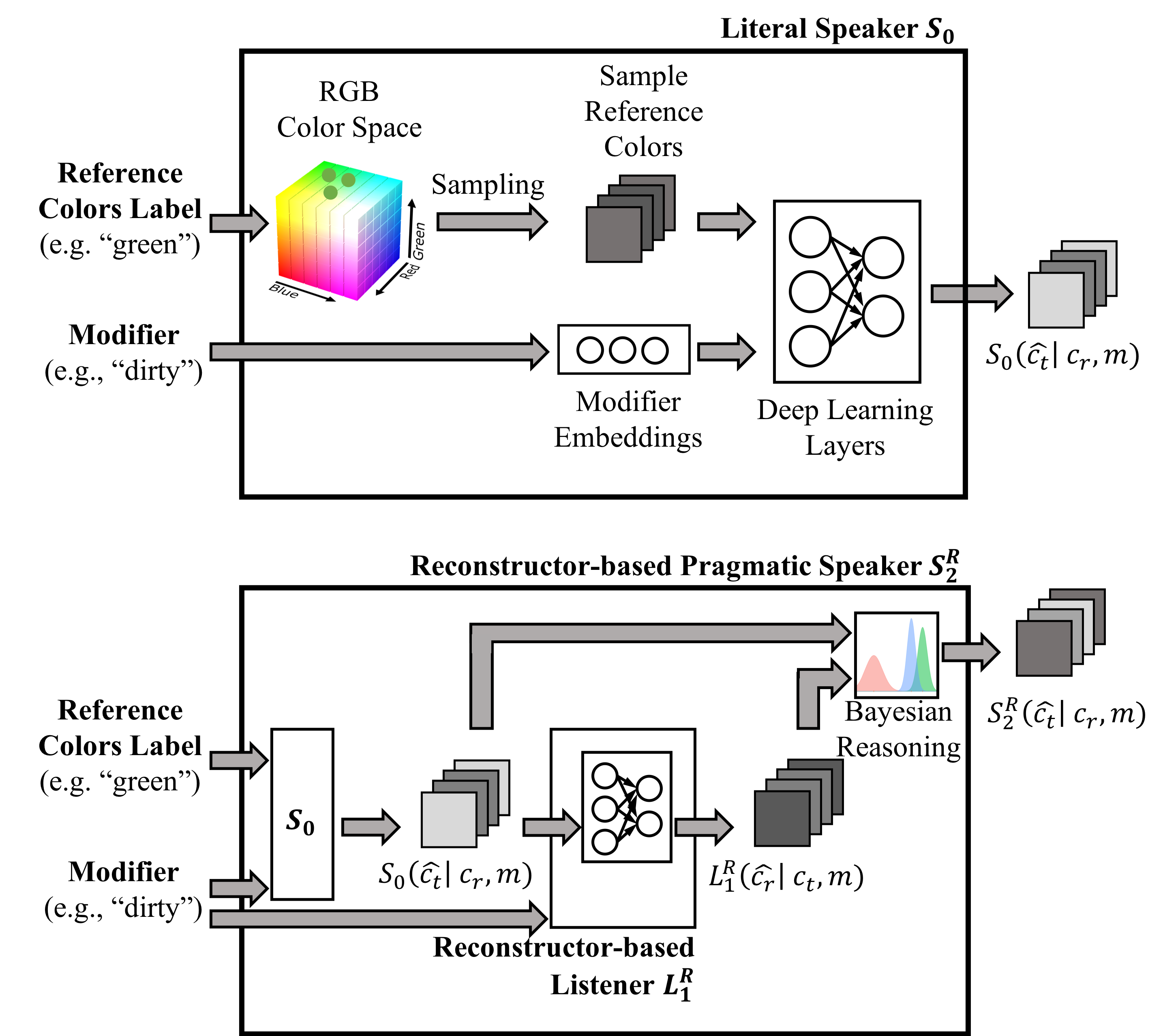}
\caption{Model architecture for our \emph{reconstructor-based pragmatic speaker} model, which consists a literal speaker model and a implicit reconstructor-based listener model.}
  \label{fig:prag-model}
\end{figure}

\section{Pragmatic Model}
The task is formulated as a conditional color generation task. Given a set of sampled RGB vectors \bm{$c_{r}$} for a reference color label and a modifier $m$, our objective is to generate a RGB vector $\hat{c}_{t}$ that is close to the gold-standard RGB vector $c_{t}$ for the target color label. 
By integrating pragmatic modeling, we focus on generating target colors that are pragmatically informative, i.e., that a potential listener can use to reconstruct the reference colors. 
%\dco{have to rephrase this previous sentence} \zen{Re: Edit this a little. I still feel this sentence is underinformative. I try to say we are not only, like deep learning way, that generate target color using modifier and reference color, but also consider the reconstruction correctness to increase informativeness.}
This process involves two main steps as shown in Fig.~\ref{fig:prag-model}. First, a \emph{literal speaker} model generates and evaluates multiple RGB vectors for a target color given the sampled RGB vectors for a reference color and a modifier. Second, a \emph{reconstructor-based pragmatic speaker} model ``reasons'' about an embedded \emph{reconstructor-based listener} that evaluates the output of the literal speaker model, and thus generates RGB vectors that best allows the listener to reconstruct the reference color RGB vector. % reevaluate the probability distribution in the previous step by considering an embedded \emph{reconstructor-based listener} model that evaluates the generated RGB vectors for the target color by reconstructing the RGB vector for the reference color from these the generated RGB vectors.
We note that the listener is better thought of as a subroutine in the pragmatic speaker model rather than a separate model, and we will discuss it as such.

\paragraph{Literal Speaker} We first describe the base \emph{literal speaker} model, 
%We build our pragmatic model on top of a \emph{literal speaker} 
$S_{0}$, which generates a probability distribution $S_{0}(\hat{c}_{t} | c_r, m)$ over target colors $\hat{c}_{t}$ given a reference color $c_r \in \bm{c_r}$ and a modifier $m$ (represented using word embeddings). To parameterize $S_{0}$, we build our \emph{literal speaker} model using the deep neural network proposed by ~\citet{winn2018lighter}. The model takes two inputs as shown in Fig.~\ref{fig:dnn}: the modifier $m$ and the RGB vector $c_{r}$ for the reference color label. An input modifier is represented by using 300-dimensional GloVe word embeddings~\cite{pennington2014glove}. We represent the modifiers as a bi-gram to account for comparatives that need \textit{``more''} (e.g. \textit{``more vibrant''}); single-word modifiers are padded with the zero vector. Both the word embeddings $m$ of modifiers and the color RGB vector $c_{r}$ are concatenated and fed into a fully connected feedforward layer with a hidden size 30:

\begin{figure}[t]
\centering
\includegraphics[width=0.9\columnwidth]{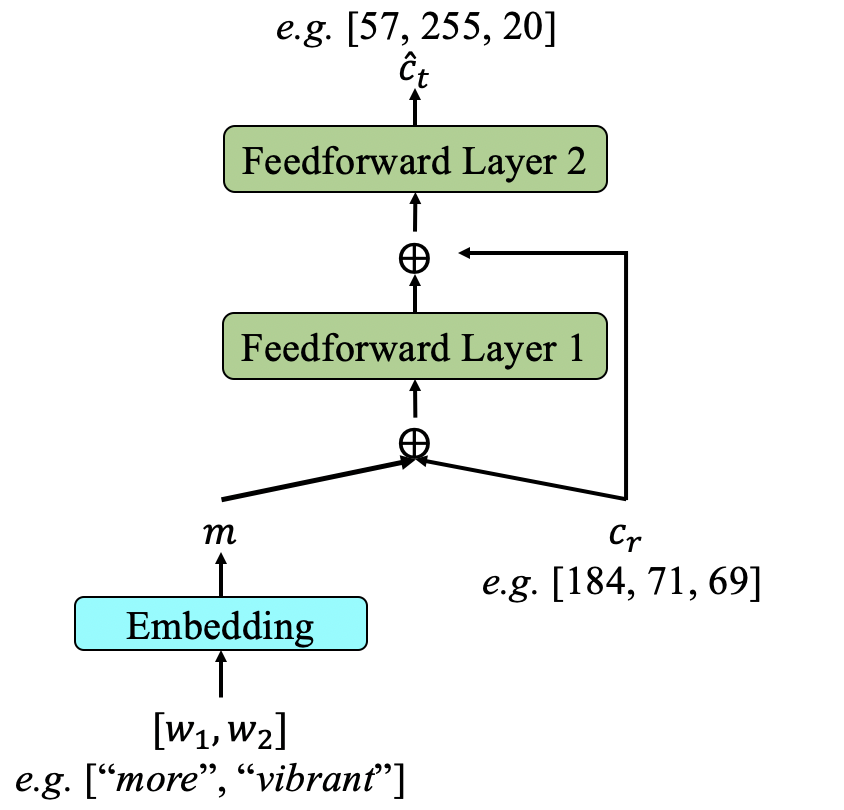}
\caption{Model architecture for our base \emph{literal speaker} $S_0$ model, which is based on the model proposed by~\citet{winn2018lighter}.}
  \label{fig:dnn}
\end{figure}

\begin{equation}    
\begin{split}
    f_{\phi}(c_{r}, m) &= \textbf{W}_{1}[c_{r}, m] + \textbf{b}_{1}
\end{split} 
\end{equation}
where $\textbf{W}_{1}$ and $\textbf{b}_{1}$ are learnt parameters. The output hidden state $o_{1}$ of the first layer is concatenated with the color RGB vector, and fed into another fully connected layer $f_{\psi}$ with a hidden size 3 to predict the RGB vector $\hat{c}_{t}$ for a target color label:
\begin{equation}    
\begin{split}
    f_{\psi}(o_{1}, m) &= \textbf{W}_{2}[o_{1}, c_{r}] + \textbf{b}_{2}\\
    \hat{c}_{t} &= f_{\psi}(o_{1}, m)
\end{split} 
\end{equation}
where $\textbf{W}_{2}$ and $\textbf{b}_{2}$ are learnt parameters. Following previous works~\cite{winn2018lighter}, our loss function two parts: (1) minimizing the cosine distance between the ``true'' color modification in RGB space, $(c_t - c_r)$, and the predicted modification $(\hat{c}_t - c_r)$. (2) minimizing the mean square error between $c_t$ and $\hat{c}_t$. Intuitively, the first part supervises the model to learn contextual meanings of modifiers, while the second part supervises the model to draw colors accurately.

%\dco{somewhere, we have to clarify the difference between the retrained WM18 and S0 in Table 2. If I understand correctly, WM18 and S0 have the same model architecture, but they differ because S0 has the additional sampling step?} \zen{Re: Yes. S0 has the sampling and evaluation by distance step. S0 will pick out the color with highest probability. I added some extra sentences below in color.}

\begin{table*}[!t]
    \centering
    \scalebox{0.95}{
    \setlength\tabcolsep{12pt}
    \begin{tabular}{@{}lcccc@{}}
    \toprule
    \multirow{2}{*}{\textbf{Test Set}} & \multicolumn{4}{c}{\textbf{Models}} \\ \cmidrule(l){2-5} 
     & \multicolumn{1}{c}{WM18} &  \multicolumn{1}{c}{HSC19-HSV$_{\emph{retrain}}$} & \multicolumn{1}{c}{$S_{0}$} & \multicolumn{1}{c}{$S_{2}^{R}$} \\ \midrule
    \multicolumn{5}{c}{Cosine Similarity (Std. Dev.); Higher = Better} \\ \midrule
    Seen Pairings (SP) & .680 (-) &  .869 (.023) & .926 (.002) & \textbf{.934 (.001)} \\
    Unseen Pairings (UP) & .680 (-) & .739 (.115) & .823 (.005) & \textbf{.834 (.003)}\\
    Unseen Ref. Color (URC) & .400 (-) & .424 (.088) & .751 (.007) & \textbf{.803 (.004)}\\
    Unseen Modifier (UM) & .410 (-) & .620 (.132) & .603 (.002) & \textbf{.690 (.002)}\\
    Fully Unseen (FUN) & -.210 (-) & .408 (.099) & .552 (.011) & \textbf{.564 (.009)}\\
    Overall (OR) & .650 (-) & .758 (.056) & .855 (.003) & \textbf{.859 (.002)}\\ \midrule
    \multicolumn{5}{c}{Delta-E Distance (Std. Dev.); Lower = Better} \\ \midrule
    Seen Pairings (SP) & 6.10 (-) & 5.58 (.022) & 5.03 (.012) & \textbf{4.79 (.011)} \\
    Unseen Pairings (UP) & 7.90 (-) & 8.00 (.082) & 5.79 (.008) & \textbf{5.72 (.006)} \\
    Unseen Ref. Color (URC) & 11.4 (-) & 20.5 (1.33) & 10.1 (.011) & \textbf{9.49 (.013)} \\
    Unseen Modifier (UM) & 10.5 (-) & 16.0 (1.28) & 13.2 (.072) & \textbf{13.2 (.066)} \\
    Fully Unseen (FUN) & 15.9 (-) & 16.4 (.665) & 15.0 (.011) & \textbf{15.0 (.006)} \\
    Overall (OR) & 6.80 (-) & 8.96 (.324) & 6.88 (.005) & \textbf{6.57 (.003)} \\ \bottomrule
    \end{tabular}
    }
    \vspace{3pt}
    \caption{Average cosine similarity score and Delta-E distance over 10 runs, of our literal speaker ($S_{0}$) and our reconstructor-based pragmatic speaker ($S_{2}^R$) in comparison to previous models: $\text{WM18}$ proposed by~\citet{winn2018lighter}, and $\text{HSC19-HSV}$ proposed by~\citet{han2019grounding}. To facilitate comparison with our models, we retrained HSC19-HSV, with more details in the main text. (-) indicates that standard deviations were not reported in the original paper. \textbf{Bolded} values indicate the best performances.}
    \label{tab:SummaryOfResults}
\end{table*}

Previous works that integrate pragmatics into deep learning solve discrimination problems---distinguishing a referent target from similar distractors~\cite{andreas2016reasoning, monroe2017colors, cohn2018pragmatically, cohn2018incremental, nie2020pragmatic}. Without the explicit presence of distractors, we use instead samples of $c_r \in \bm{c_r}$ to serve as distractors. For $\bm{c_r}$, we randomly sample $n$ RGB vectors for each color label.\footnote{We randomly sample 100 RGB vectors for a particular color, and use the mean RGB values of these 100 RGB vectors as the RGB values for one sample. We iterate this process $n$ times.} We pass each sample through the deep learning layers, to generate multiple target colors. As a result, the model predicts $n$ modified RGB vectors $\hat{c}_{t} \in \bm{\hat{c}_{t}}$ for a target color label. In our experiments, we set $n$ to 10.

% For $\bm{c_r}$, we randomly sample $n$ RGB vectors for each color label.\footnote{We randomly sample 100 RGB vectors for a particular color, and use the mean RGB values of these 100 RGB vectors as the RGB values for one sample. We iteration this process $n$ times.} As a result, the model outputs $n$ predicted modified RGB vectors $\hat{c}_{t} \in \bm{\hat{c}_{t}}$. In our experiments, we set $n$ to 10.

To evaluate generated $\hat{c}_{t}$ via sampling, we formulate the probability distribution as $S_{0}(\hat{c}_{t} | c_r, m)$, by using the distances between RGB vectors in RGB space:
\begin{equation}    
\begin{split}
    S_{0}(\hat{c}_{t} | c_r, m) &\propto \frac{e^{\Delta(\hat{c}_{t}, \bar{\bm{c_r}})}}{\sum_{\hat{c}_{t}^{'}} e^{\Delta(\hat{c}_{t}^{'}, \bar{\bm{c_r}})}} \label{eqn:s0}
\end{split} 
\end{equation}
where $\bar{\bm{c_r}}$ is the mean RGB vector associated with the reference color label. We propose that modified target color is expected to diverge from the reference color in the color space as in~\citet{winn2018lighter}. We use Delta-E 2000 in CIELAB color space as $\Delta(\cdot)$ in Eqn.~\ref{eqn:s0} and Eqn.~\ref{eqn:lr} (See Sec.~\ref{Sec:results} for the description and other distance metrics). The RGB vector with the highest probability is considered as the final output for $S_{0}$. We use Delta-E 2000 in CIELAB which is also commonly used in previous works for evaluating model performances in color generation~\cite{winn2018lighter} (See Sec.~\ref{sec:eval} for the definitions). Note that this distance measurements can be replaced by other metrics.  

Thus to summarize, our literal speaker model $S_{0}$ takes the base deep learning model proposed by WM18, adds a sampling procedure to produce distractors, and optimizes for target colors that maximize distance in color space.  Though adding the sampling procedure is quite intuitive, previous deep learning models suffers from the lack of capability to incorporate the richness of the color representations.

\begin{figure*}[tb]
    \centering
    \includegraphics[width=0.8\textwidth]{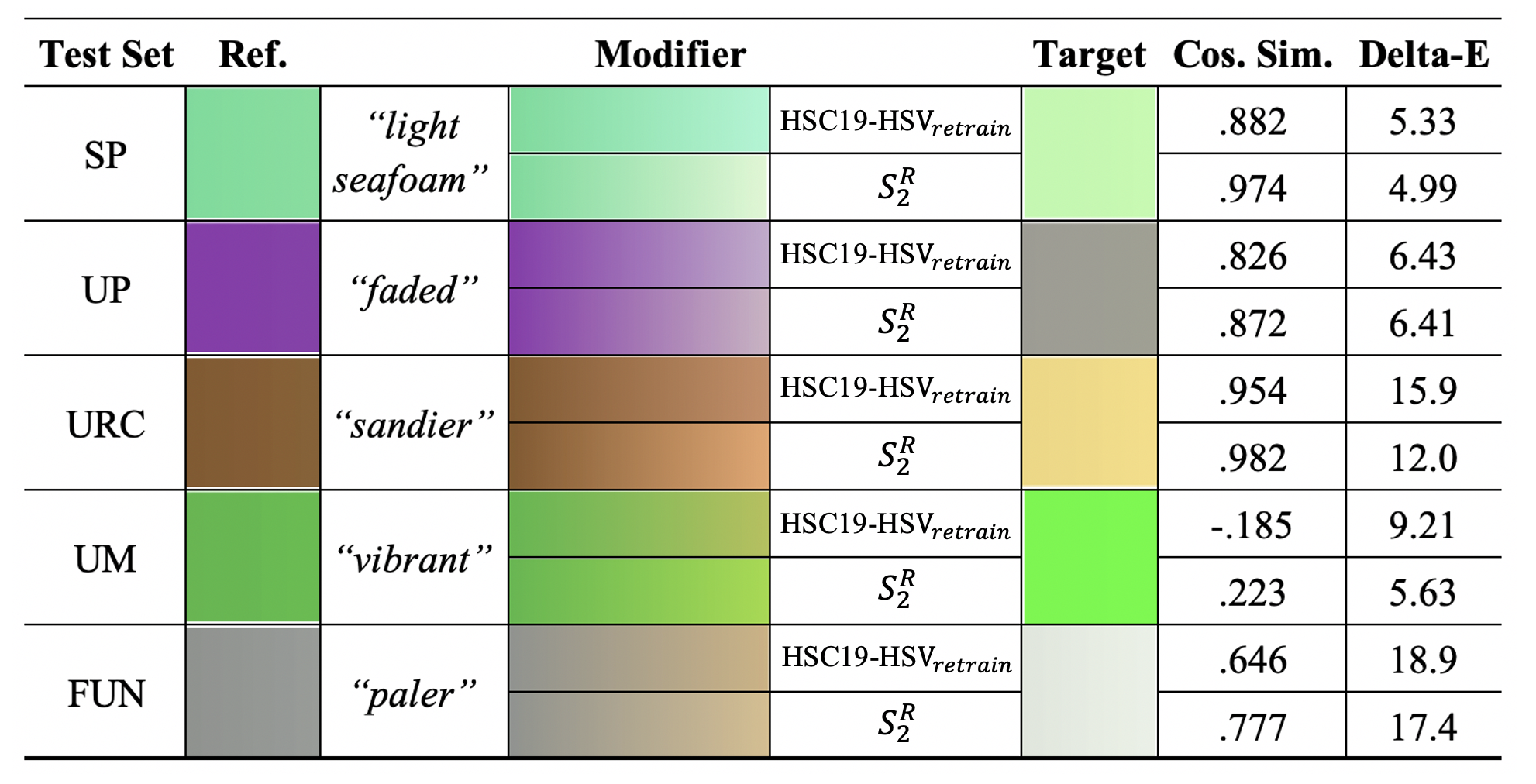}
    \caption{Examples of predictions for our retrained model~\cite{han2019grounding} $\text{HSC19-HSV}$, and our \emph{reconstructor-based pragmatic speaker} model $S_{2}^{R}$. The Test Sets include: Seen Pairings (SP), Unseen Pairings (UP), Unseen Reference Color (URC), Unseen Modifiers (UM), and Fully Unseen (FUN). For each example, we include reference color (Ref.), predicted modifier, true target color, and results based on our evaluate metrics.}
    \label{fig:error}
\end{figure*}

\paragraph{Reconstructor-Based Pragmatic Speaker} To increase informativeness of our outputs from $S_{0}$, we further propose a ``listener'' that maximizes the probability of reconstructing inputs from the outputs of our \emph{literal speaker} model. Previous works denote this as a \emph{reconstructor-based listener}  $L_{1}^{R}$~\cite{duvsek2016sequence, fried2017unified, shen2019pragmatically}. The listener produces a probability distribution $L_{1}^{R}( c_r | \hat{c}_{t}, m)$ by reconstructing $c_r$ given a $\hat{c}_{t}$ and $m$. We parameterize $L_{1}^{R}$ using a separate neural network which has the same model architecture as in $S_{0}$ trained with the objective of predicting $c_r$ given $c_{t}$ and word embeddings for $m$. 

Similar to Equation \ref{eqn:s0}, we formulate the probability distribution for $L_{1}^{R}( c_r | \hat{c}_{t}, m)$ by considering reconstruction error using the distance between RGB vectors in the RGB space:
% To generate the probability distribution for $S_{0}(\hat{c}_{t} | c_r, m)$, we use distance between RGB vectors in the RGB space:
% the reconstructed RGB vector for a reference color label given a predicted RGB vector for a target color label $\hat{c}_{t}$ and a modifier $m$. We parameterize $L^{R}$ using the same model architecture as in $S_{0}$. 
% over the RGB vector of a reference color given $\hat{c}_{t}$. For $L^{R}$, we use the same model architecture as $S_{0}$, but retraining with target colors and modifiers as the inputs, and reference colors $c_r$ as the output:
\begin{equation}    
\begin{split}
    L_{1}^{R}( c_r | \hat{c}_{t}, m) &\propto \left( \frac{e^{\Delta(\hat{c}_{r}, \bar{\bm{c_r}})}}{\sum_{\hat{c}_{r}^{'}} e^{\Delta(\hat{c}_{r}^{'}, \bar{\bm{c_r}})}} \right)^{-1} \label{eqn:lr}
\end{split} 
\end{equation}
where $\hat{c}_{r}$ is the reconstructed RGB vector for a reference color label. Unlike Equation \ref{eqn:s0}, we have the probability proportional to the \emph{inverse} of the distance function as we want the reconstructed reference color to be close to the true reference color. 

Now, we introduce our pragmatic model, the \emph{reconstructor-based pragamatic speaker} model, which combines the \emph{literal speaker} and the \emph{reconstructor-based listener} (Fig.~\ref{fig:prag-model}). This pragmatic speaker model has as ``subroutines'' that it can call, the literal speaker and the reconstructor-based listener. It ``reasons'' about the listener, and chooses the most informative colors that the listener can use to reconstruct the original reference colors (which it does by reasoning about the literal speaker).
Formally, the reconstructor-based pragmatic speaker generates a probability distribution over $\hat{c}_{t}$, weighting both the $L_{1}^{R}$ and $S_{0}$ terms:
\begin{equation}    
\begin{split}
   S_{2}^{R}(\hat{c}_{t} | c_r, m) &= L_{1}^{R}( c_r | \hat{c}_{t}, m)^{\lambda} \cdot S_{0}(\hat{c}_{t} | c_r, m)^{1 - \lambda} \label{eqn:s1r}
\end{split}
\end{equation}
where $\lambda$ is a pragmatic reasoning parameter that controls how much the model optimizes for discriminative outputs, following ~\cite{monroe2017colors}. The predicted RGB vector with the highest probability in Eqn.~\ref{eqn:s1r} is our pragmatic prediction for $\hat{c}_{t}$. We found an optimal value of $\lambda$ at 0.33 using grid search by evaluating with the \emph{Validation} set.
%For $n$, we have found that can be relatively small; in our experiments, it is set to 10. $\lambda$ find to be optimized at 0.33 using grid search. 

% We predict that our pragmatic speaker outperforms existing models in generalizing unseen data given its integration with generative sampling and pragmatic inference.

\section{Experiment Setup}\label{sec:eval}

\paragraph{Evaluation:} We evaluate our models under 5 different test sets as in~\citet{winn2018lighter, han2019grounding}: (1) 271 Seen Pairings (SP): The triple (reference color label, modifier, target color label) has been seen in the training set. (2) 29 Unseen Pairings (UP): The reference color label and modifier have been seen in the training set, but not their pairing (i.e., they appeared separately). (3) 63 Unseen Reference Color (URC): reference color label is not present in the training set, but modifier is. (4) 41 Unseen Modifiers (UM): modifier is not present in the training set, but the reference color label is. (5) Fully Unseen (FUN): Neither reference color label or modifier has been seen in the training data. (6) Overall (OR): All samples in the testing set. As mentioned in Sec.~\ref{Sec:data}, we use a different set of RGB vectors in the test sets for a color label that has been seen during training to ensure our model is not overfitting with the training set. We averaged our performance results over 10 runs with distinct random seeds. 

We use two different metrics to evaluate our models, following~\citet{winn2018lighter, han2019grounding}: (1) Cosine similarity scores between the two difference vectors $(c_t - c_r)$ and $(\hat{c}_t - c_r)$; Higher scores correspond to better performance. (2) Delta-E distance in CIELAB color space between $c_t$ and $\hat{c}_t$ (see~\citet{winn2018lighter, han2019grounding} for discussions). Delta-E is a non-uniformity metric for measuring color differences, which was first presented as the Euclidean Distance in CIELAB color space~\cite{mclaren1976xiii}. ~\citet{luo2001development} present the latest and most accurate CIE color difference metrics, Delta-E 2000, which improves the original formula by taking into account weighting factors and fixing inaccuracies in lightness. Lower Delta-E 2000 values correspond to better performance.

\paragraph{Model configuration:} To train our models, we use a standard laptop with 2.9 GHz Intel Core i7 and 16 GB 2133 MHz LPDDR3 with GPU training disabled. With this computing infrastructure, it takes about 10 minutes to train all of our models, where each model is trained with 500 epochs. Our \emph{literal speaker} model $S_{0}$ contains 18,222 trainable parameters which is the same as \emph{reconstructor-based listener}  model $L_{1}^{R}$. Our reconstructor-based listener pragmatic speaker $S_{2}^R$ contains 36,445 trainable parameters.

\section{Results} \label{Sec:results}
Table~\ref{tab:SummaryOfResults} shows test results, in comparison with previous best performing models. We retrain the best performing model $\text{HSC19-HSV}$~\cite{han2019grounding}
with our evaluation pipeline to ensure fairness\footnote{We note that in the original HSC19 paper, they computed the cosine similarity between vectors in HSV space and RGB space. We think this could have been an inadvertent mistake, and so we retrained HSC19-HSV and used cosine similarities between RGB vectors, which results in some differences in our table. %For HSC19-HSV, we use cosine similarity between vectors in the RGB color space instead of between vectors vectors in HSV space and RGB space as in the original paper which we considered as an much less meaningful measurement~\cite{han2019grounding}.
Details can be found in their public code repository~\url{https://github.com/HanXudong/GLoM}.}.

Compared to the both models, our literal model $S_{0}$ achieves better performance across most test cases while maintains similar performance in others as shown in the Table~\ref{tab:SummaryOfResults}. This is expected given the fact that $S_{0}$ maximizes information gain via additional evaluating with sampling procedure. Our pragmatic model $S_{2}^{R}$ further increases performances across all \emph{Test} sets by outperforming our literal model $S_{0}$ with significant gains, which is expected given the model adds pragmatic informativeness. For all the test sets, $S_{2}^{R}$ exceeds the state-of-the-art (SOTA) deep learning based models across both evaluation metrics. For more difficult test cases where one of the input is unseen, our models perform extremely well. For instance, our $S_{2}^{R}$ model has a 95\% absolute increase in cosine similarity score for the Unseen Reference Color (URC) test set. For the Fully Unseen (FUN) test set where both the reference color and the modifier are unseen during training, our $S_{2}^{R}$ still shows the ability to generalize to new test samples with a 40\% increase in cosine similarity score comparing to previous models who perform poorly with unseen data.

The performance gain from our pragmatic models suggests that contextual descriptions of colors might be pragmatically generated by participants. For example, when participants are asked to describe a color \crule[dirty_green]{0.4cm}{0.4cm}, % (``dirty green'')
previous studies have found people rely on using adjective modifiers (e.g., ``\emph{dirty}'') with a reference to a base color (e.g., ``\emph{green}'')~\cite{lassiter2017adjectival}. The pragmatic approach offers an additional step: when participants are selecting a base color implicitly, they may also be \emph{de}-modifying the color showed to them.
%\dco{I think we can add more here. we have some space, so we should discuss our results in more detail}

\paragraph{Error Analysis: } In Fig.~\ref{fig:error}, we provide selected illustrative examples from each of the five \emph{Test} sets. We discuss the two cases from the UM and FUN test sets. For the UM example, the reference color is ``\emph{green}'' and the modifier is ``\emph{vibrant}''. Since in our training, the word embedding for ``\emph{vibrant}'' is unseen, the non-pragmatic (literal) model results in a negative cosine similarity. 

On the other hand, our pragmatic model $S_{2}^{R}$ is capable of making more accurate predictions when presented with previously unseen 
%{\color{red}highly uninformative} \dco{should this be: ``unseen''? why is this ``highly uninformative''} \zen{Re: Yes, it is unseen. I was trying to say for unseen data, it is kind of less informative to the model as the model has not seen data like this during training. If seen, then it can gain more informativeness?} 
test cases. This is expected as, by performing additional recursive evaluation, the pragmatic model is able to choose more ``informative'' target colors. 
%\dco{technically, the recursion doesn't ``add'' information...} \zen{Re: Right. Maximize information gain?}
This is consistent across all \emph{Test} sets where we have unseen test examples. For FUN test set, both $S_{0}$ and $S_{2}^{R}$ models make poor predictions. We expect that for unseen modifiers, the predictions should be based on modifiers with similar meanings. For example, ``\emph{paler}'' should be similar to some seen modifiers, e.g. ``\emph{whiten}'', ``\emph{faded}''. However, the predicted modification is more close to ``\emph{sandier}'' in this case. As described in~\citet{mrkvsic2016counter}, this may related to closeness in the word embedding space for these modifiers.

\begin{figure}[t]
\centering
\includegraphics[width=1\columnwidth]{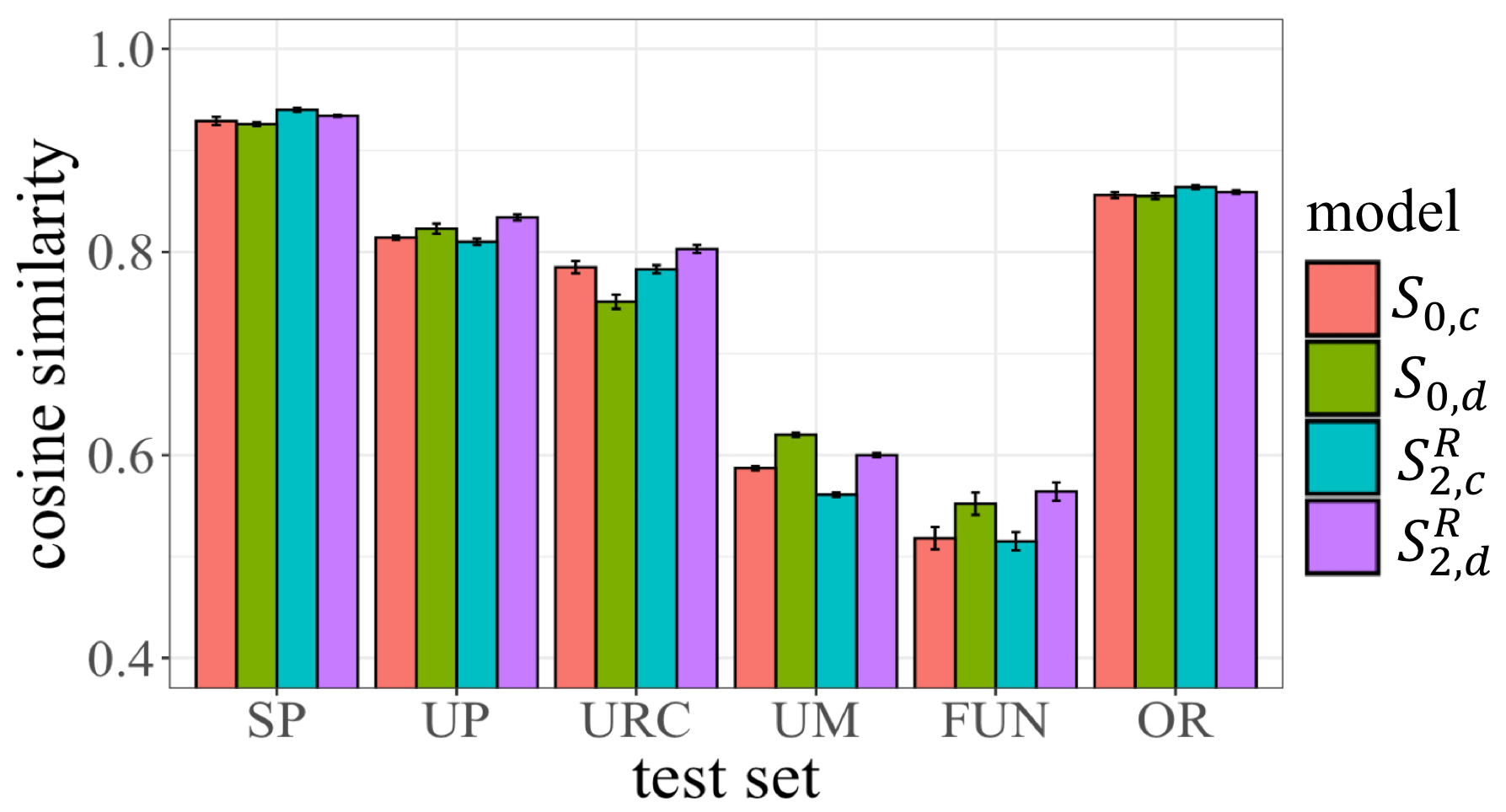}
\caption{Test result for each \emph{Test} set evaluated by cosine similarity with different distance-based probability estimations over 10 runs. Error bars represents standard deviation of the cosine similarity. }
  \label{fig:dis-cosine}
\end{figure}

\begin{figure}[t]
\centering
\includegraphics[width=1\columnwidth]{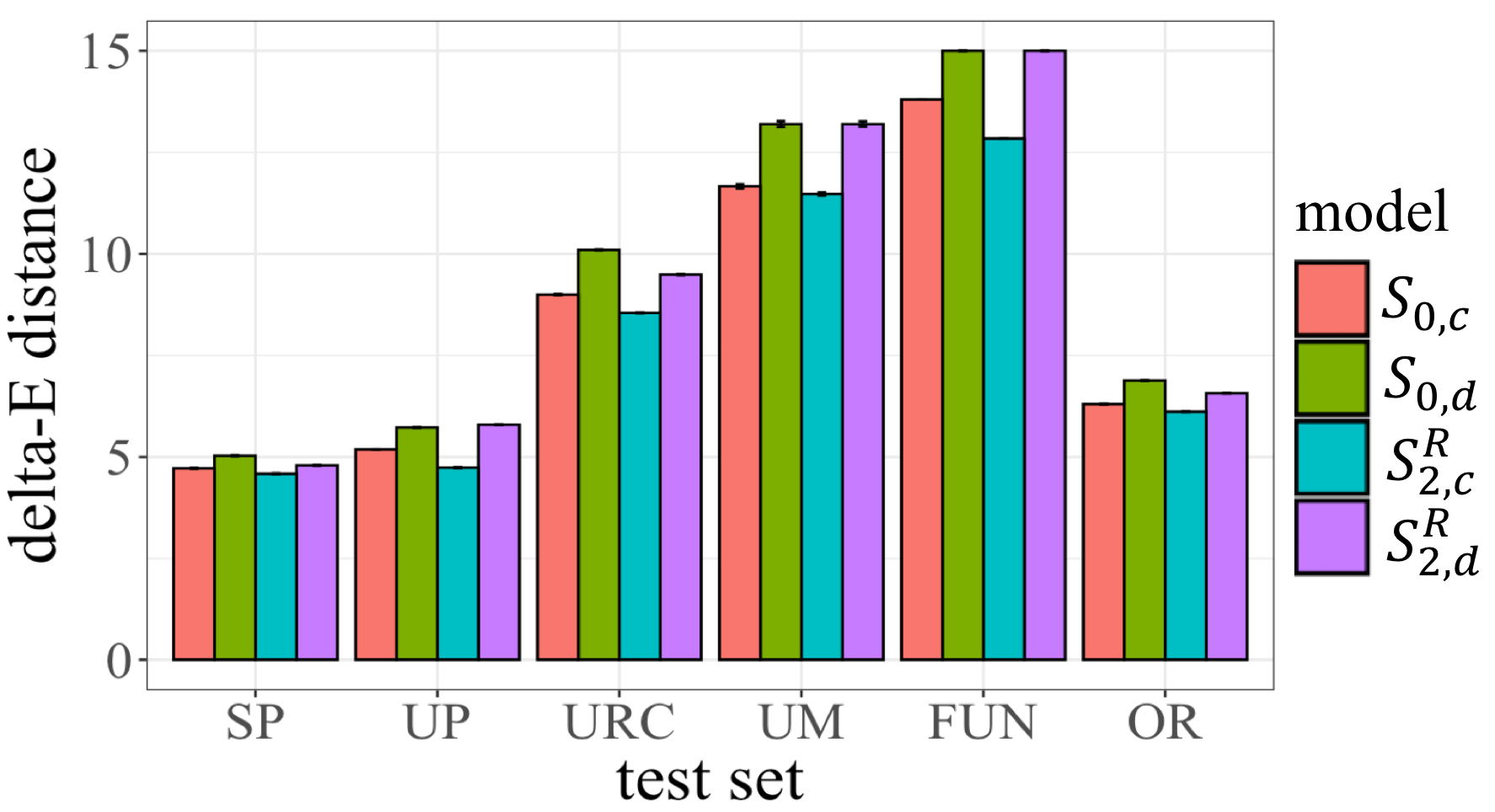}
\caption{Test result for each \emph{Test} set evaluated by Delta-E 2000 distance~\cite{luo2001development} in CIELAB color space~\cite{mclaren1976xiii} with different distance-based probability estimations over 10 runs. Error bars represents standard deviation of the Delta-E 2000 distance.}
  \label{fig:dis-deltae}
\end{figure}

\paragraph{Distance Metrics: } As mentioned, we use Delta-E 2000 distance in the CIELAB color space as $\Delta(\cdot)$ in Eqn.~\ref{eqn:s0} and Eqn.~\ref{eqn:lr} by default. Additionally, we use cosine distance in the RGB color space as our distance metrics to formulate our distance-based probability estimations. 
%\dco{can we standardize cosine distance or cosine similarity? I'm a bit confused after reading this and the captions for Fig. 2 and 3... and also ``$_{cos}$''} \zen{Re: Addressed. Change some places to using distance not similarity. I changed the footnote as well.} 
We denote $S_{0, \text{c}}$ and $S_{2, c}^{R}$ for cosine distance based estimation, and $S_{0, d}$ and $S_{2, d}^{R}$ for Delta-E 2000 based estimation. Results are shown in Fig.~\ref{fig:dis-cosine} and Fig.~\ref{fig:dis-deltae}. Our result suggests that Delta-E 2000 based models perform better in most of test cases. This result corroborates with previous works suggesting Delta-E 2000 based distance in CIELAB color space is more accurate in describing nuanced difference between colors~\cite{luo2001development}. As shown in Fig.~\ref{fig:dis-cosine} and Fig.~\ref{fig:dis-deltae}, \emph{reconstructor-based pragmatic speaker} models are consistently outperforming \emph{literal speaker} models across multiple \emph{Test} sets.

\section{Conclusion and Future Work}

In this paper, we propose a novel \emph{reconstructor-based pragmatic speaker} model, and apply it to a color generation task that requires extensive grounding of contextual modifiers in color space. Our base \emph{literal speaker} model, with no pragmatics and which adapts previous deep learning models with distractor sampling, performs respectably on modifiers and colors that it was trained on, but performs poorly when testing on previously unseen modifiers and colors. Our pragmatic speaker model reasons about a listener that tries to guess the speaker's input, and hence provides pragmatically informative utterances (colors). The output of the pragmatic speaker model shows consistent improvements in performance over the base literal speaker model across multiple testing conditions, unseen test words (e.g. unseen colors). 
We note that in this case (and in many other datasets as well), the data was generated by people---perhaps people may be applying some pragmatic reasoning when labeling colors. We speculate that this might both (i) explain the performance of the pragmatic speaker model, and (ii) justify the pragmatic assumptions we make in building our model. 
This hypothesis would need to be further confirmed with additional human experiments, but overall we are hopeful that this bodes well for this general endeavour of augmenting deep learning approaches with pragmatics to build better models of natural language understanding.

%In this paper, we propose a novel \emph{reconstructor-based pragmatic speaker} model, and apply it to a color generation task that requires extensive grounding of contextual modifiers. For contextual color generation, our results show that our \emph{literal speaker} models which is built on previous deep learning models with distractor sampling, while strong in performance results, perform poorly when testing against unseen data. Our pragmatic models encourages predictions that can be used to reconstruct their inputs, and these informative predictions show consistent improvements over various testing conditions, especially for unseen test words (e.g. unseen colors). We speculate that this result may provide insights about how the data was generated (i.e., perhaps people apply pragmatic reasoning when labeling colors?), which further adds to the utility of augmenting ``traditional'' deep learning approaches with pragmatics to understand human-generated text. 

Our pragmatic pipeline provides another successful example of applying pragmatic modelings in traditional machine learning generation tasks, and the first to do so with a simple ``image'' (color) output. Future work could explore merging this approach with deep learning methods in other more complex contextual image generation tasks, such as generating images from natural language captions. 

\bibliography{emnlp2020}
\bibliographystyle{acl_natbib}

\end{document}